\pdfoutput=1

\documentclass[11pt]{article}

\usepackage[final]{acl}

\usepackage{times}
\usepackage{latexsym}
\usepackage{hyperref}

\usepackage{amsmath}
\usepackage{amssymb, amsfonts}
\DeclareMathOperator*{\argmax}{arg\,max}

\usepackage{bbm}
\usepackage[T1]{fontenc}

\usepackage[utf8]{inputenc}

\usepackage{microtype}

\usepackage{inconsolata}

\usepackage{graphicx}

\title{In-Domain African Languages Translation Using LLMs \\ and Multi-armed Bandits}
\author{Pratik Rakesh Singh, Kritarth Prasad, Mohammadi Zaki \and Pankaj Wasnik \\
        Media Analysis Group, Sony Research India\\
        \texttt{\{pratik.singh, kritarth.prasad, mohammadi.zaki, pankaj.wasnik\}@sony.com}}

\begin{document}
\maketitle
\begin{abstract}


Neural Machine Translation (NMT) systems face significant challenges when working with low-resource languages, particularly in domain adaptation tasks. These difficulties arise due to limited training data and suboptimal model generalization, As a result, selecting an optimal model for translation is crucial for achieving strong performance on in-domain data, particularly in scenarios where fine-tuning is not feasible or practical. In this paper, we investigate strategies for selecting the most suitable NMT model for a given domain using bandit-based algorithms, including Upper Confidence Bound, Linear UCB, Neural Linear Bandit, and Thompson Sampling. Our method effectively addresses the resource constraints by facilitating optimal model selection with high confidence. We evaluate the approach across three African languages and domains, demonstrating its robustness and effectiveness in both scenarios where target data is available and where it is absent.

\end{abstract}


\section{Introduction}
Advancements in multilingual machine translation models have significantly expanded language coverage, enabling translations even for low-resource languages. These models have also demonstrated strong performance in general domains, such as News, Movies, and more \cite{barrault-etal-2020-findings-first} \cite{saunders2022domainadaptationmultidomainadaptation}. Additionally, with the rise of large language models, methods like few-shot learning and in-context learning have shown notable improvements in domain adaptation tasks \cite{garcia2023unreasonableeffectivenessfewshotlearning} \cite{aycock-bawden-2024-topic}. Despite these advancements, the performance of these models remains highly dependent on the quality and scope of pre-training data as well as the model size, particularly for low-resource languages. It is common for a model to perform well in one domain but struggle in another, which presents a significant challenge in the selection of the most suitable NMT system for a given task.

\begin{figure}
    \centering
    \includegraphics[width=1\linewidth]{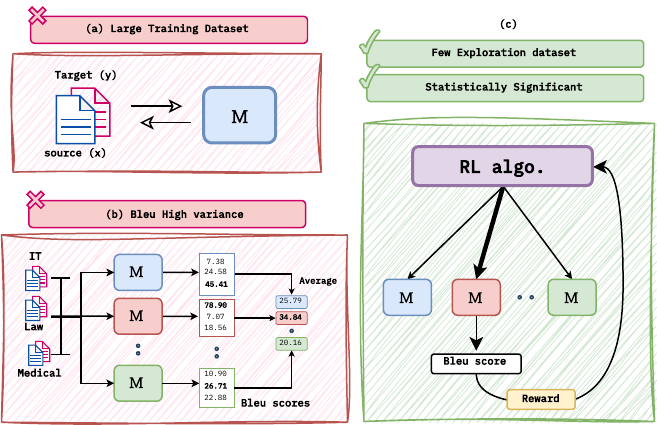}
    \caption{Motivation for Reinforcement Learning for model selection in machine translation: (a) Using a large dataset for training may be inefficient or impractical for low-resource settings, (b) BLEU scores vary significantly across domains, making model selection unreliable, (c) Reinforcement learning enables efficient model selection with fewer data and statistical significance.}
    \label{fig:motivation}
\end{figure}

A commonly used approach for domain adaptation in Neural Machine Translation (NMT) tasks is fine-tuning NMT models on in-domain data using various strategies \cite{chu-wang-2018-survey}. As shown in Figure \ref{fig:motivation} (a), this approach faces significant challenges, particularly in low-resource settings where in-domain data is scarce. Moreover, fine-tuning often leads to a degradation in performance on general-domain data due to the issue of catastrophic forgetting \cite{thompson-etal-2019-overcoming}, further complicating the task of maintaining robust model performance across different domains.

Selection-based approaches have gained significant attention in recent Neural Machine Translation (NMT) systems, where the task is to identify the best possible model from a given set. A widely adopted method for this is the use of a Selection Block (SB) \cite{salazar-etal-2020-masked} \cite{liu-liu-2021-simcls}, which reranks models based on the specific task at hand. In recent years, reinforcement learning (RL)-based approaches have emerged as promising techniques for optimizing the selection of these models \cite{prasad2025fastermachinetranslationensembling}. However, a limitation of many of these approaches is their reliance on large amounts of data to demonstrate performance gains over individual NMT systems.

One potential approach is to select the best model, which is trained on a general-domain dataset, and assume it will perform well on the in-domain dataset without the need for fine-tuning. Typically, one would evaluate models on the test dataset using common machine translation metrics, such as BLEU, to determine the best-performing model. However, in resource-constrained settings, these metrics can exhibit high variance, and there is often limited control over the statistical significance of the observed differences. This issue is illustrated in Figure \ref{fig:motivation} (b), where the mean of these metrics might provide a misleading impression of the best NMT system. In such cases, a few outlier examples could skew the BLEU score, leading to the wrong selection of the model. Moreover, a single evaluation does not capture the full variability in system performance, particularly when working with a small validation set. This underscores the need for model selection methods that not only choose the best NMT systems but also provide a statistical basis for the selection process, thereby mitigating the risks of misleading conclusions based on limited data.

To address the above challenges of model selection for Domain adaptation in resource-constrained settings, one possible approach could be to estimate the most optimal NMT system using fewer data samples, thereby reducing the reliance on large datasets. This can be achieved through the use of bandit-based algorithms \cite{zhou2016surveycontextualmultiarmedbandits} \cite{9185782}, which allow for efficient exploration and exploitation of model performance, facilitating the identification of the best-performing system for the given domain with minimal data. As shown in Figure \ref{fig:motivation} (c), by leveraging these techniques, it is possible to make more informed decisions about model selection, even when In-domain data availability is limited, ensuring effective performance in low-resource scenarios. Our key contributions are summarized as follows: 
\begin{itemize}
    \item We propose a bandit-based approach to estimate optimal systems for a domain in a resource-constraint setting.
    \item We evaluate our approach on English to multiple African languages in multiple domains and report the performance of the popular bandit algorithms when applied to domain-specific model selection task.
\end{itemize}

\section{Related works and Motivation}
{\bfseries Domain Adaptation} in Neural Machine Translation (NMT) refers to methods aimed at adjusting translation models trained on general-domain data to perform effectively in specific target domains with distinctive characteristics \cite{saunders2022}. Effective domain adaptation typically addresses data scarcity and domain mismatch \cite{challenges_nmt} issues through data-centric and model-centric approaches. Data-centric strategies include back-translation using monolingual target data \cite{poncelas2019adaptationmachinetranslationmodels, jin2020simplebaselinesemisuperviseddomain}, forward-translation and self-learning \cite{chinea-rios-etal-2017-adapting}, and synthetic data generation via noise introduction or lexicon-based methods \cite{vaibhav-etal-2019-improving, hu-etal-2019-domain-adaptation, peng2020dictionarybaseddataaugmentationcrossdomain, zhang-etal-2022-iterative}. Model-centric approaches introduce domain-specific parameters or modules like domain tagging, embedding manipulation, adapter-based methods, and pointer-generators leveraging dictionaries \cite{kobus2017domaincontrolneuralmachine, stergiadis-etal-2021-multi, pham-etal-2019-generic, bapna-firat-2019-simple, chen2021}.

In low-resource scenarios, approaches such as data augmentation through bilingual lexicon-based replacements, transfer learning, and pretrained multilingual models have been employed \cite{nag2020incorporatingbilingualdictionarieslow, liu-etal-2021-continual}. However, despite significant progress in both DA and low-resource NMT, domain adaptation techniques remain underexplored and challenging specifically for low-resource languages, where often the only available parallel data are very limited \cite{siddhant20221000languagesmultilingualmachine, NMT_survey}.\\

\noindent \textbf{Selection-Based Approach} Recently, various selection methods have been introduced prior to the fusion step in multi-agent candidate selection. Significant research has focused on summarization tasks, including training reranking models based on evaluation metrics \cite{ravaut2023summarerankermultitaskmixtureofexpertsreranking}, employing contrastive learning for effective candidate ranking \cite{liu-liu-2021-simcls}, and utilizing pairwise ranking methods to directly compare candidate summaries \cite{jiang-etal-2023-llm}. In the field of neural machine translation (NMT), recent studies by \cite{prasad2025fastermachinetranslationensembling} have explored model selection strategies using a DQN-based approach. However, all these selection methods require substantial amounts of parallel data for effective training. Notably, limited research has investigated contextual bandit \cite{contextual_bandit} approaches, which require significantly less data, to generalize agent selection based on provided context in low-resource machine translation scenarios.\\

\begin{figure}
    \centering
    \includegraphics[width=1\linewidth]{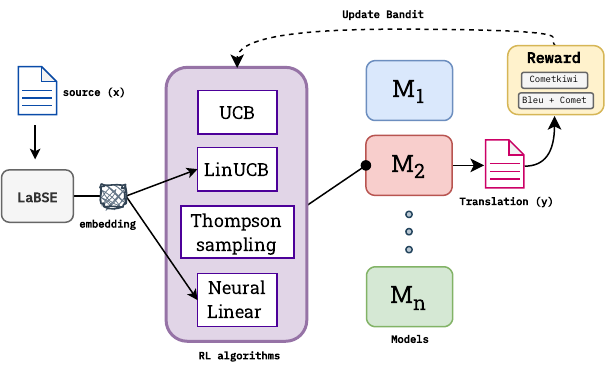}
    \caption{Block diagram of the proposed bandit-based model selection strategy.}
    \label{fig:architecture}
\end{figure}
\noindent {\bfseries Motivation for Bandit based approach} A major challenge in Low-Resource Machine Translation (LRMT) is the scarcity of high-quality training datasets. This issue is further compounded in domain-specific translation, where the data becomes even more limited. While general-domain NMT systems exhibit reasonable performance across a broad range of tasks, their efficacy significantly fluctuates across different domains and languages. To mitigate this variability and identify the most effective NMT model for a given task with limited data, an optimal selection strategy is essential. This strategy must not only consider the available training data but also provide statistically-backed confidence in the model's selection. 

Bandit-based approaches have been widely explored in recommendation systems, where recommendations are generated based on past interactions with users \cite{SILVA2022116669}. This methodology is well-suited for selecting optimal NMT systems in scenarios where only a small in-domain dataset is available, utilizing an appropriate reward function for NMT performance \cite{boursier2024surveymultiplayerbandits} \cite{nguyen2017reinforcementlearningbanditneural}. Furthermore, reference-less reward mechanisms offer a promising avenue for applying these bandit-based methods in target-free domain-specific machine translation tasks, as demonstrated by recent works \cite{Obuchowski_Klaudel_Frąckowski_Krajna_Badyra_Czubenko_Kowalczuk_2024}.

\section{Methodology}
As previously discussed, selecting the best model from a pool by evaluating a subset of data and then applying it to the entire test set is both computationally expensive and unreliable. Determining the necessary sample size to ensure the optimality of the chosen model becomes extremely important in such cases. Hence we take a more principled way to dynamically choose the machine translation model on-the-fly by treating the model selection process as a multi-armed bandit problem. We explore popular bandit algorithms designed for regret minimization, which, under mild theoretical assumptions, are proven to achieve (near-)optimal cumulative rewards over time. Below, we provide a brief overview of our methodology, as illustrated in Figure~\ref{fig:architecture}. 

Each source sentence $x$ is passed through a Language-agnostic BERT Sentence Encoder (LaBSE) to obtain a feature vector which we denote by overloading $x\in \mathbb{R}^d$. This vector $x$ acts as the context vector in the contextual bandit algorithms considered in this work. The MT system pool act as the arms $\{M_1, M_2,\ldots, M_n\}$ in our multi-armed bandit setup. Once the arm is chosen by the MAB algorithm, the corresponding MT system is chosen to translate the source sentence $x$ to obtain $y$ in the target language. Next, a reward is generated depending on $x$, $y$ and whether a reference gold translation is available (see the next section for detail on `reward') to obtain a scalar $r$.

Next, we provide a brief explanation of the arm selection strategy and update rules for each of the bandit algorithms we explore.

\begin{itemize}
    \item \textbf{Upper Confidence Bound (UCB):} UCB \cite{Auer2002} relies on the principle of Optimism in the Face of Uncertainty (OFU). It selects the arm that maximizes an `upper confidence bound' of its estimated reward.
    \begin{align*}
        &\text{\textbf{Select arm}:}\\
        &a_t:= \argmax\limits_{a\in \{M_1,M_2,\ldots, M_n\}} \left(\hat{\mu}_a(t)+\alpha\sqrt{\frac{\log t}{N_a(t)}} \right).\\
        &\text{\textbf{Update} empirical means of all arms.}
    \end{align*}
where $\hat{\mu}_a(t)$ is the empirical reward obtained from pulling arm $a$ till round $t$, $N_a(t)$ is the number of times arm $a$ is pulled till round $t$, and $\alpha$ is a confidence parameter.
    \item \textbf{Thompson Sampling (TS):} Thompson Sampling \cite{thompsonsampling} is a Bayesian approach where we maintain a posterior distribution over each arm's expected reward and sample from it. In particular, in our case we maintain a Beta distribution over each arm's reward which has two parameters $\alpha, \beta$ which are initially set to 0. The arm selection and parameter update rules are as follows:
    \begin{align*}
        &\text{\textbf{Select arm}:}\\
        &a_t:= \argmax\limits_{a\in \{M_1,M_2,\ldots, M_n\}} \theta_a\sim P(\cdot|\alpha_a, \beta_a)\\
        &\text{\textbf{Update}:}\\
        &\alpha_a\gets \alpha_a + r, 
        \beta_a\gets \beta_a + 1-r
    \end{align*}

    \item \textbf{Linear UCB (LinUCB):} LinUCB \cite{linucb, oful} extends UCB to contextual bandits, assuming that rewards follow a linear function of the context/feature vector $x_t\in \mathbb{R}^d$ as explained before . In particular, we make the following assumption on the reward function that $\forall t\geq1,$ $r:=x^T\theta_a + noise$ for all arms $a$.
    \begin{align*}
        &\text{\textbf{Select arm}:}\\
        &a_t:= \argmax\limits_{a\in \{M_1,M_2,\ldots, M_n\}} \left( x_t^T\hat{\theta}_a + \alpha\sqrt{x_t^TA_a^{-1}x_t}        \right)
    \end{align*}
where, $A_a:=\sum\limits_{s=1}^t \mathbbm{1}\{a_s==a\} x_{s}x_{s}^T$, $b_a:=\sum\limits_{s=1}^t \mathbbm{1}\{a_s==a\} r_sx_{s}$ and $\hat{\theta}_a:=A_a^{-1}b_a$ is the Least Squares estimate of the true parameter $\theta_a$ of arm $a$.
    \item \textbf{Neural LinUCB (NL):} Neural LinUCB \cite{neurallinucb} is a deep-learning extension of LinUCB, replacing the linear model with a neural network that maps features to a latent representation before applying LinUCB. In particular we replace the context vector $x$ by a neural network $f(x:w)$ parameterized by $w$. The arm selection strategy and the update rule remain the same as in LinUCB with $x$ replaced with $f(x;w)$.
\end{itemize}

\noindent {\bfseries Rewards:} The rewards serve as the primary signal in bandit-based settings, guiding both the learning process and decision-making of the algorithms. The main objective in a Multi-Armed Bandit (MAB) problem is to maximize rewards by balancing exploration and exploitation.

In Neural Machine Translation (NMT), model performance is typically evaluated using standard metrics such as BLEU and COMET. These metrics are particularly crucial for assessing how well a model translates within a specific domain. BLEU measures how accurately the model translates domain-specific vocabulary, while COMET evaluates the semantic similarity of the model’s output to the reference translation within the given domain. Using these metrics, we consider two types of reward signals as follows:
\begin{itemize}
    \item {\bfseries When parallel data is present: } When we have source along with the reference (gold) translation, we consider a combination of BLEU \citep{sacrebleu} and a reference-based comet as shown below. Note that both the BLEU and comet scores have been normalized to lie between [0,1].\\
    \begin{align*}
    \begin{split}
           \text{Reward} &= \lambda \cdot\text{BLEU} + (1 - \lambda) \cdot{\text{COMET}}.
           \end{split}
    \end{align*}
Here $\lambda$ is a hyperparameter in [0,1]. In our experiments, we find that $\lambda=0.4$ achieves the best results in our case.

    \item {\bfseries Target-free scenario: } When only the source sentence is present, and the target is absent, which is typically the case in low-resource languages, especially in domain translation task,  we use a reference-less MT metric like CometKiwi \citep{rei-etal-2022-cometkiwi} (normalized between [0,1]), as the reward signal.
\end{itemize}
The $\lambda$ is a controllability parameter that enables us to control the influence of metrics on reward, and reward\_norm is the normalization value to normalize the metrics.

\begin{table*}[ht]
    \centering
    \resizebox{\textwidth}{!}{
    \begin{tabular}{|c|c|c|c|c|c|c|c|c|c|} \hline 
         Datasets&  News Igbo&  News Yoruba&  News Swahili&  Movies Igbo&  Movies Yoruba&  Movies Swahili&  Religious Igbo&  Religious Yoruba& Religious Swahili\\ \hline 
         Aya101&  12.98&  5.20&  23.53&  7.48&  3.847&  23.98&  19.03&  11.53& 5.04\\ \hline 
         Gemma2 9B&  8.17&  3.55&  24.24&  5.21&  1.97&  25.07&  9.48&  2.12& 3.9\\ \hline 
         Llama 3.1 8B&  4.59&  3.28&  17.24&  2.85&  1.09&  10.90&  6.79&  3.43& 2.25\\ \hline 
         Madlad&  6.91&  1.11&  8.92&  7.087&  1.19&  24.47&  3.19&  1.36& 32.36\\ \hline 
         NLLB&  19.73&  9.67&  27.57&  9.60&  12.90&  \textbf{30.78}&  34.72&  14.96& 28.01\\ \hline 
         UCB&  \textbf{19.83}&  9.539&  \textbf{28.275}&  9.6&  12.90&  \textbf{30.78}&  34.72&  14.96& 32.36\\ \hline 
         TS&  19.48&  \textbf{9.74}&  26.95&  9.36&  12.88&  27.29&  34.34&  14.34& \textbf{32.54}\\ \hline 
         LinUCB&  19.73&  9.54&  27.80&  \textbf{9.80}&  12.90&  29.9&  \textbf{34.8}&  13.7& 32.45\\ \hline
 NL& 19.74& 9.67& 27.57& 9.24& \textbf{13.12}& 24.37& 34.72& \textbf{15.67}&32.39\\\hline
    \end{tabular}
    }
    \caption{Performance on BLEU metrics when Parallel data is present.}
    \label{tab:seed_b+c_table}
\end{table*}

\begin{table*}[ht]
    \centering
    \resizebox{\textwidth}{!}{
    \begin{tabular}{|c|c|c|c|c|c|c|c|c|c|} \hline 
         Datasets&  News Igbo&  News Yoruba&  News Swahili&  Movies Igbo&  Movies Yoruba&  Movies Swahili&  Religious Igbo&  Religious Yoruba& Religious Swahili\\ \hline 
         Aya101&  12.98&  5.20&  23.53&  7.48&  3.847&  23.98&  19.03&  11.53& 5.04\\ \hline 
         Gemma2 9B&  8.17&  3.55&  24.24&  5.21&  1.97&  25.07&  9.48&  2.12& 3.9\\ \hline 
         Llama 3.1 8B&  4.59&  3.28&  17.24&  2.85&  1.09&  10.90&  6.79&  3.43& 2.25\\ \hline 
         Madlad&  6.91&  1.11&  8.92&  7.087&  1.19&  24.47&  3.19&  1.36& 32.36\\ \hline 
         NLLB&  19.73&  9.67&  27.57&  \textbf{9.60}&  \textbf{12.90}&  \textbf{30.78}&  \textbf{34.72}&  \textbf{14.96}& 28.01\\ \hline 
         UCB&  \textbf{19.83}&  9.53&  \textbf{28.27}&  7.48&  \textbf{12.90}&  \textbf{30.78}&  \textbf{34.72}&  \textbf{14.96}& 28.01\\ \hline 
         TS&  19.48&  \textbf{9.74}&  26.95&  9.33&  12.89&  27.95&  34.34&  14.34& 32.54\\ \hline 
         LinUCB&  18.5&  9.67&  27.5&  8.9&  \textbf{12.90}&  29.7&  34.54&  11.51& \textbf{32.67}\\ \hline
 NL& 19.73& 9.67& 27.57& 7.48& \textbf{12.90}& 23.98& \textbf{34.72}& 11.53&32.36\\\hline
    \end{tabular}
    }
    \caption{Performance on BLEU metrics in Target-Free scenario}
    \label{tab:target_free}
\end{table*}

\section{Experimental Setup}
{\bfseries Datasets and Evaluation metrics: } For our experiments, we utilize parallel datasets for English-to-African language translation, focusing on three African languages: English-to-Yoruba (en-yo), English-to-Swahili (en-sw) and English-to-Igbo (en-ig). We sample 1,000 parallel samples for validation (seed data for model convergence) and testing each. The datasets span three domains: News, Movies, and Religious texts. 
\begin{itemize}
    \item News Domain: We use the Lafand-MT dataset \cite{adelani-etal-2022-thousand}, which contains parallel data for English-to-16 African languages, gathered from various news corpora.
    \item Movies Domain: We leverage the OpenSubtitles dataset \cite{lison-tiedemann-2016-opensubtitles2016}, which includes parallel translations of dialogues from various movies and TV shows. This domain is essential for capturing informal language usage and conversational nuances in translations.
    \item Religious Texts Domain: We compile a dataset from various sources, including CCAligned \cite{elkishky2020ccalignedmassivecollectioncrosslingual} and Tanzil (available at \url{https://tanzil.net/}), which contains Quran translations in multiple languages. This domain is particularly valuable for translating formal, religious content.
\end{itemize}
To assess the performance of the models, we rely on BLEU \cite{10.3115/1073083.1073135}, a widely accepted metric in machine translation. BLEU effectively measures the degree of overlap between the model-generated translations and reference translations, capturing the adequacy of domain-specific vocabulary translation, which is specifically effective for In-Domain Translations\\
{\bfseries Models: } We test the effectiveness of our approach by using baselines, which also act as arms for bandits. The models used are a Mixture of LLMs and Foundational models like Aya101 \cite{ustun2024ayamodelinstructionfinetuned}, NLLB200 3.3B (NLLB) \cite{nllbteam2022languageleftbehindscaling}, Madlad400 10B (Madlad) \cite{kudugunta2023madlad400}, Gemma2 9B \cite{gemmateam2024gemma2improvingopen} and Llama3.1 8B \cite{grattafiori2024llama3herdmodels} all the model used for experiments are in its based or pre-trained state. \\
{\bfseries Choice of Hyper-parameters:} For evaluation of our proposed bandit-based strategies, we use four popular arm-selection bandit algorithms that are UCB, LinUCB, Neural Bandit, Thompson Sampling as discussed in detail in Sec. 3. The values of the hyper-parameters specific to each bandit strategy are given in Table~\ref{tab:hyperparameters}. The selection of Hyper-parameters was done based on hit and trail method, where initial few sentences of validations were used for convergence of algorithm and the algorithm was tested on rest of the remaining sentences.

\begin{table}[]
\resizebox{.5\textwidth}{!}{
\begin{tabular}{|l|l|l|}
\hline
Algorithm         & Parameter                                                                             & Value                                                       \\ \hline
UCB               & $\lambda$                         & [0.4-0.6] \\ \hline
Thompson Sampling & Prior Distribution                                                                    & Beta(0,0)                                                   \\ \hline
LinUCB            & \begin{tabular}[c]{@{}l@{}}$\alpha$\\ $\lambda$\end{tabular} & \begin{tabular}[c]{@{}l@{}}1.5\\ 0.4\end{tabular}     \\ \hline
Neural LinUCB     & Arm model network                                                                     & \begin{tabular}[c]{@{}l@{}}2 layer MLP\\ with 50 neurons each\end{tabular}               \\ \hline
\end{tabular}
}
\caption{Values of hyper-parameters used in our experiments.}
\label{tab:hyperparameters}
\end{table}

\section{Results}
{\bfseries When parallel data is present:} The hyper-parameters are tuned on the validation set and freezed for all the algorithms. The Bandit-based models explored the reward that is a weighted summation of BLEU and reference-based Comet on the exploration set (Section 3 Rewards), the performance is evaluated using 1000 test samples per domain and language. As shown in Table \ref{tab:seed_b+c_table}, most of the Bandit-based approaches, when selecting the optimal arm, either perform on par with or surpass the best possible NMT model, demonstrating the effectiveness of our proposed method. On average, UCB (Upper Confidence Bound) achieves superior performance across all languages and domains, outperforming the best model, NLLB, by an average improvement of 2.68\% in BLEU score. Additionally, in specific cases—such as Neural Linear Bandit (NL) in the Movies domain for Yoruba, UCB in the News domain for Igbo and Swahili, and Thompson Sampling (TS) in the News domain for Yoruba, Religious, and Swahili the Bandit-based algorithms surpass the performance of the best possible NMT model. This suggests that these algorithms can occasionally select alternative NMT systems, resulting in slight but notable improvements in translation quality. In summary, the results indicate that Bandit-based approaches can effectively identify the best-performing NMT models, even with very small training sets, highlighting the robustness and utility of our proposed training strategy. \\

\noindent{\bfseries Results on Target-free Scenario: } In this experiment, we explore the scenario where target translations for the in-domain dataset are unavailable, a common challenge in low-resource language settings. Such cases can be addressed using reference-less rewards (Section 3 Rewards), specifically leveraging CometKiwi-based metrics for NMT evaluation. The exploration of the Bandit-based models follows the same setup as discussed previously, with testing performed on 1000 samples. As shown in Table \ref{tab:target_free}, the Bandit-based approaches successfully identify the best arms for translations even in the absence of target translations for reward generation. Among the various Bandit algorithms, UCB performs the best, followed by LinUCB and NL. Notably, in some instances, the Bandit-based selection slightly outperforms the individual best models, underscoring the flexibility of our approach. This demonstrates that our method can effectively be applied to model selection in target-free domain translation task where reference translations are not available.

\section{Conclusion}
In this paper, we presented a bandit-based approach for selecting the most suitable NMT model for domain adaptation, particularly in low-resource settings. Our method effectively balances exploration and exploitation by leveraging strategies such as Upper Confidence Bound, Linear UCB, Neural Linear Bandit, and Thompson Sampling, enabling optimal model selection with high confidence. Experimental results across multiple African languages and domains confirm the robustness of our approach, demonstrating its effectiveness both in the presence and absence of target domain data. Our findings highlight the potential of bandit-based methods to improve NMT performance in resource-constrained environments, paving the way for a more efficient and adaptive model selection process.
\bibliography{africanlp_CR}
\newpage

\end{document}